\documentclass[10pt,twocolumn]{article}

\usepackage[letterpaper,margin=0.72in]{geometry}
\usepackage{authblk}
\usepackage{booktabs}
\usepackage{amsmath,amssymb}
\usepackage{graphicx}
\usepackage{microtype}
\usepackage[numbers,sort&compress]{natbib}
\usepackage{tikz}
\usetikzlibrary{arrows.meta,positioning}
\usepackage{balance}
\usepackage[hidelinks]{hyperref}

\title{\textbf{Observational Policy Ranking for SMB Financial Guidance from Multi-Action Accounting Logs}}
\author[1]{Shrutendra Harsola}
\author[1]{Vignesh Subrahmaniam}
\author[1]{Vikas Raturi}
\author[1]{Kamalika Das}
\author[1]{Xiang Gao}
\author[1]{Kratika Gupta}
\author[1]{Ruocheng Guo}
\author[1]{Padmaja Jonnalagedda}
\author[1]{Ananya Pramod}
\author[1]{Sricharan Kumar}
\affil[1]{Foresight-AI, Intuit}
\date{August 2026}

\hypersetup{
  pdftitle={Observational Policy Ranking for SMB Financial Guidance from Multi-Action Accounting Logs},
  pdfauthor={Shrutendra Harsola; Vignesh Subrahmaniam; Vikas Raturi; Kamalika Das; Xiang Gao; Kratika Gupta; Ruocheng Guo; Padmaja Jonnalagedda; Ananya Pramod; Sricharan Kumar},
  pdfkeywords={observational policy ranking, financial decision support, multi-action logs, model-assisted comparison}
}

\begin{document}
\raggedbottom
\maketitle

\begin{abstract}
Small and medium-sized businesses need timely financial guidance, yet historical accounting logs record self-selected and often co-occurring business changes rather than randomized recommendations. We formulate this setting as observational policy ranking: from pre-decision financial information, a policy selects one of 34 ledger-derived business-change categories for a target financial KPI. Using 85,078 company-month observations from 7,505 firms, we introduce Covariate-Adjusted Residual Policy Learning (CAR-PL), an action-wise R-learner that operates directly on multi-hot logs and regularizes selection by observational support. We compare CAR-PL with an uplift T-Learner, a conservative contextual value model, a zero-shot LLM, and non-personalized references on company-disjoint held-out firms under a shared model-assisted scoring rule. CAR-PL has the highest Gross Profit point estimate (0.084), the T-Learner has the highest Revenue point estimate (0.085), and the contextual value model has the highest Quick Ratio point estimate (0.062). CAR-PL and the T-Learner are not statistically separated on either growth KPI in matched company-clustered comparisons, while CAR-PL selects 33--34 categories and produces less concentrated selections across the catalog. Outcome-model-only scoring retains the same KPI-level point-estimate leader or top pair, and category rankings remain similar when the all-zero treatment reference is replaced by the most common training co-action pattern. These findings support objective-specific ranking of SMB financial guidance from multi-action accounting logs.
\end{abstract}



\section{Introduction}
Small and medium-sized businesses (SMBs) often need advice before financial problems become acute, yet expert advisory capacity is costly and difficult to scale. Randomized consulting programs show that external guidance can change business practices and outcomes, but such programs are expensive to deliver broadly \cite{bruhn2018}. Accounting systems provide a complementary source of evidence: detailed financial histories together with the business changes that firms subsequently make. Converting these histories into candidate guidance requires learning from observational, multi-action records rather than from randomized recommendation labels.

We formulate the task as \emph{observational policy ranking}. At a prospective decision boundary, a policy observes information available through month $t{+}0$, selects one category from a fixed 34-category catalog, and targets a subsequent financial KPI. Historical categories are ledger-derived business-change proxies detected during month $t{+}1$; outcomes are measured from $t{+}2$ through $t{+}12$. The action month is excluded from both state and outcome, producing a clean prospective timeline for every company-month decision.

Three properties make the task difficult: (1) category occurrence is strongly state dependent, creating systematic selection; (2) multiple categories can occur in the same month, complicating focal credit assignment; and (3) observational support varies sharply across the 34 categories, making the policy argmax sensitive to rare-category noise. The learner must therefore estimate conditional contrasts and select stably across unevenly observed candidates. Within each KPI, a credible comparison holds the action space, retained test rows, outcome definition, and scoring rule fixed across policy families.

We compare a conservative contextual value model, an uplift T-Learner, and CAR-PL, a covariate-adjusted action-wise R-learner with support shrinkage. A zero-shot LLM, a modal constant policy, and a frequency-weighted random reference provide complementary comparisons. Within each KPI, every method uses the same catalog and retained company-disjoint test rows. A shared model-assisted focal-action score combines an outcome-model contrast with propensity-weighted residual evidence.

The point-estimate pattern is KPI-specific: CAR-PL has the highest Gross Profit point estimate, the T-Learner the highest Revenue point estimate, and the contextual value model the highest Quick Ratio point estimate. CAR-PL and the T-Learner are not statistically separated on either growth KPI in matched company-clustered comparisons. On those KPIs, they agree on only 12.8--14.2\% of company-month states, while CAR-PL selects 33--34 categories. Thus, comparable aggregate scores can arise from markedly different state-to-category mappings, making recommendation concentration a substantive dimension of policy comparison.

Our contributions are threefold. First, we formulate prospective policy ranking from multi-action accounting logs, with a fixed pre-decision, action, and outcome timeline and company-disjoint evaluation across 7,505 firms. Second, we introduce CAR-PL, which combines action-wise R-learning with support shrinkage to stabilize one-of-34 selection while retaining the original multi-hot observations. Third, we benchmark learned, zero-shot, and non-personalized policies under a single frozen scorer and report KPI-dependent point-estimate leaders, robustness across scoring variants, and sharply different recommendation-concentration profiles.

\section{Related Work}
\textbf{Observational policy learning.} Policy-learning methods use observational data to optimize treatment rules under explicit adjustment and overlap assumptions. Athey and Wager develop efficient observational policy learning for binary treatment, while Zhou, Athey, and Wager extend doubly robust policy learning to multiple actions \cite{athey2021policy,zhou2018offline}. Our setting contributes a large industrial application in which the logged representation is multi-hot and the policy emits one interpretable category. Work on multiple versions of treatment provides the conceptual basis for defining focal effects when observed treatment categories occur in heterogeneous bundles \cite{vanderweele2013versions}.

\textbf{Heterogeneous effects and support-aware selection.} Meta-learners estimate conditional treated-control contrasts by reducing heterogeneous-effect estimation to supervised learning \cite{kunzel2019metalearners}. R-learning residualizes both outcome and treatment against pre-treatment covariates and estimates the remaining conditional effect signal \cite{nie2021quasi}. CAR-PL builds on this construction with action-wise heads and a support-based shrinkage step designed for argmax selection over many categories. This complements uncertainty-aware and pessimistic policy-learning methods that explicitly penalize weakly supported decisions \cite{jin2025pessimism}.

\textbf{Offline value learning and financial decisions.} Offline reinforcement learning addresses policy optimization from fixed logs and must control overestimation on unsupported actions \cite{fujimoto2019bcq,levine2020offline}. CQL implements this principle through a conservative value penalty \cite{kumar2020cql}; we adapt its one-step objective as a contextual value baseline. Machine learning is increasingly used in credit and financial decision systems \cite{khandani2010consumer,fuster2022predictably}, and language models are being explored for recommendation and financial advice \cite{dai2023llm_rec,fieberg2025llm_finance}. The zero-shot policy in our study measures the ranking induced by pretrained business knowledge without fitting to historical outcomes.

\textbf{SMB guidance as decision support.} Financial recommendation differs from conventional product recommendation because the output is an operational business change and the outcome unfolds over subsequent accounting periods. Prior work studies consulting interventions for small firms \cite{bruhn2018}, algorithmic credit decisions \cite{khandani2010consumer,fuster2022predictably}, and financial advice from language models \cite{fieberg2025llm_finance}. Our study brings these strands together in a repeated company-month setting with prospective timing and a common interpretable catalog. It targets the combination left open by these strands: multi-hot observational actions, a single interpretable policy output, and KPI-specific evaluation on unseen firms.

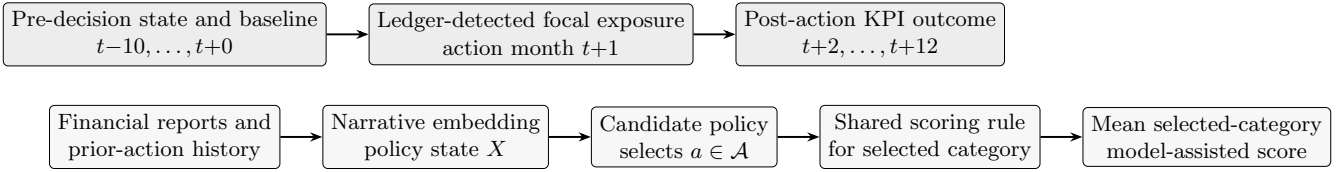
\begin{figure*}[t]
\centering
\resizebox{0.98\textwidth}{!}{%
\begin{tikzpicture}[
 box/.style={draw,rounded corners=2pt,minimum height=0.72cm,minimum width=2.55cm,align=center,font=\small,fill=black!3},
 tbox/.style={draw,rounded corners=2pt,minimum height=0.68cm,minimum width=3.1cm,align=center,font=\small,fill=black!7},
 arr/.style={-{Stealth[length=5pt]},thick},
 node distance=0.55cm and 0.60cm]
 \node[tbox] (pre) {Pre-decision state and baseline\\$t{-}10,\ldots,t{+}0$};
 \node[tbox,right=of pre] (act) {Ledger-detected focal exposure\\action month $t{+}1$};
 \node[tbox,right=of act] (post) {Post-action KPI outcome\\$t{+}2,\ldots,t{+}12$};
 \draw[arr] (pre)--(act);
 \draw[arr] (act)--(post);

 \node[box,below=0.55cm of pre] (reports) {Financial reports and\\prior-action history};
 \node[box,right=of reports] (state) {Narrative embedding\\policy state $X$};
 \node[box,right=of state] (policy) {Candidate policy\\selects $a\in\mathcal A$};
 \node[box,right=of policy] (eval) {Shared scoring rule\\for selected category};
 \node[box,right=of eval] (score) {Mean selected-category\\model-assisted score};
 \draw[arr] (reports)--(state);
 \draw[arr] (state)--(policy);
 \draw[arr] (policy)--(eval);
 \draw[arr] (eval)--(score);
\end{tikzpicture}%
}
\caption{Prospective timing and shared comparison pipeline. No action-month or post-action information enters a policy state. Every policy emits one of the same 34 categories, and within each KPI the same scoring rule is applied to the selected category on the same retained test rows.}
\label{fig:pipeline}
\end{figure*}

\section{Data and Task}
\subsection{Decision Timing, State, and Data Partitions}
Let $\mathcal A=\{1,\ldots,34\}$ denote the supported policy catalog and $\mathcal X$ the policy-state space. Each observation is a company-month pair $i$ with policy state $X_i$ and action-month exposure vector $\mathbf T_i\in\{0,1\}^{34}$. The corpus contains 85,078 observations from 7,505 firms, with a median of 12 decision months per firm. Figure~\ref{fig:pipeline} shows the fixed timing convention. The policy state and reward baseline use information through $t{+}0$; business changes are detected during $t{+}1$; and monthly KPI values from $t{+}2$ through $t{+}12$ determine the reward. The action month is excluded. Observations with an empty prior or outcome window are removed before KPI-specific denominator filters are applied.

A frozen, versioned reporting pipeline first derives structured facts and diagnostics from multiple pre-decision financial reports, including KPI trajectories, product, customer, and vendor concentration, receivables and payables, cash-flow and debt trends, and detected anomalies. Multiple LLM calls then synthesize these intermediate summaries into the final natural-language state document, which also contains month-by-month historical KPI tables. This staged construction accommodates long report histories without passing the complete raw report set to a single model call. The pipeline was fixed before policy training, receives neither action labels nor post-decision outcomes, and excludes action-month transactions. These are substantial state documents (median approximately 23,000 characters), preserving both high-level interpretation and detailed historical trajectories. The Qwen3-Embedding-0.6B model encodes each document into 768 dimensions \cite{qwen3embedding2025}. A 44-dimensional binary history block records prior-month activity from the full detector vocabulary. The ten labels below the policy-output support threshold serve only as lagged state covariates and are never eligible policy outputs. Concatenation produces the policy state $X_i\in\mathcal X\subseteq\mathbb R^{812}$, which policy learners use. Records are de-identified before modeling, processed in the approved analytics environment, and reported only in aggregate.

Firm identifiers are assigned by a deterministic pseudorandom hash, so the split is outcome independent and reproducible. The hash reserves held-out test firms, corresponding to approximately 10\% of retained rows, and divides the remaining development firms into training and validation partitions. Training rows fit policy parameters and nuisance functions. Validation rows select model configurations, early stopping, and CAR-PL's shrinkage strength. All partitions, nuisance cross-fitting folds, and early-stopping splits are defined at the firm level, so all months of a firm remain in one partition. The test set is used only for the reported policy scores and confidence intervals. Table~\ref{tab:data-summary} reports training and validation jointly as the development population.

\begin{table*}[t]
\centering
\small
\setlength{\tabcolsep}{7pt}
\caption{KPI-specific sample sizes after reward construction and filtering. Row counts are company-month observations. Development combines the company-disjoint training and validation partitions; the final column reports unique firms represented after KPI-specific filtering.}
\label{tab:data-summary}
\begin{tabular}{lrrrr}
\toprule
KPI & Total rows & Development rows & Held-out test rows & Unique firms \\
\midrule
Gross Profit & 78,091 & 70,670 & 7,421 & 7,104 \\
Revenue & 80,388 & 72,755 & 7,633 & 7,191 \\
Quick Ratio & 79,806 & 72,154 & 7,652 & 7,072 \\
\bottomrule
\end{tabular}
\end{table*}

\subsection{Action Catalog}
The extraction layer produces 44 raw ledger-derived labels. An outcome-independent support rule retains 34 labels with at least 100 treated and 100 control observations in training; these define the common policy catalog. Table~\ref{tab:domains} groups them into ten operational domains. Representative categories include increasing advertising spend, canceling subscriptions, substituting vendors, increasing training spend, securing financing, requiring upfront deposits, switching payment processors, and adopting time tracking. Directional changes are represented as separate categories where applicable. The shared vocabulary provides an interpretable action layer for all policy families.

\begin{table*}[t]
\centering
\footnotesize
\setlength{\tabcolsep}{5pt}
\caption{Ten semantic catalog domains with representative categories.}
\label{tab:domains}
\begin{tabular}{p{2.55cm}p{4.45cm}p{2.55cm}p{4.45cm}}
\toprule
Domain & Illustrative categories & Domain & Illustrative categories \\
\midrule
Revenue/Sales & Sales-team incentives & Marketing & Increase advertising spend \\
Channels/Products & Expand to a new sales channel & Subscriptions & Cancel subscriptions \\
Vendor/Purchasing & Substitute or consolidate vendors & Cost Reduction & Reduce bank or insurance fees \\
Payroll/Contractors & Increase training spend & Cash Flow/Financing & Secure financing or require deposits \\
Operations/Tools & Switch processors or adopt time tracking & Accounting/\allowbreak Compliance & Accounting workflow and payment timing \\
\bottomrule
\end{tabular}
\end{table*}

The extraction layer is rule based. It maps action-month ledger evidence to categories using account-code movements, vendor or payment-processor appearance and disappearance, recurring-payment patterns, double-entry signatures, and changes relative to a trailing historical baseline. Detectors use action-month ledger activity but not the later KPI windows that define the policy outcome. Definitions are fixed before policy training and shared across KPIs. The evidence families cover changes in spending, vendors, financing structure, recurring payments, and operating tools, giving the catalog a consistent accounting interpretation across firms. Historical labels are ledger-derived business-change proxies; for category $a$, $T_{i,a}=1$ means that the corresponding detector fired in the action month. The representation is multi-hot because several categories may be detected together.

In a separate precision-oriented audit, a sample of 1,000 positive company-month/category detections was reviewed against the underlying ledger evidence by a judge LLM calibrated against human labels. The resulting audited precision was 95\%. This positive-only audit estimates detection precision and does not estimate recall. At inference, the same vocabulary is presented as candidate guidance, and every policy is masked to the 34 supported categories before test-set scoring. The support rule is outcome-independent and shared across KPIs.

\subsection{KPI Rewards}
We train and evaluate a separate policy for Gross Profit, Revenue, and Quick Ratio. The reward compares the same 11 calendar months before and after the action month, which aligns seasonality while preserving prospective timing. Let $\kappa^k_{i,\ell}$ denote the realized value of KPI $k$ for observation $i$ at relative month $\ell$, and let $G_k(\kappa^k_{i,a:b})$ aggregate a window by summation for Gross Profit and Revenue and by averaging for Quick Ratio. All three KPIs use the normalized post-versus-pre change
\begin{equation}
 r_i^k=\frac{G_k(\kappa^k_{i,2:12})-G_k(\kappa^k_{i,-10:0})}
 {|G_k(\kappa^k_{i,-10:0})|}.
 \label{eq:reward}
\end{equation}
Rows with undefined or numerically unstable denominators are removed before modeling. Across all filtering steps, the retained Gross Profit, Revenue, and Quick Ratio samples are 91.8\%, 94.5\%, and 93.8\% of the corpus, respectively (Table~\ref{tab:data-summary}). The raw change is capped at $|r_i^k|\le 50$ and winsorized at the training-set 5th and 95th percentiles. We denote the resulting modeling reward by $R_i^k$. Because each KPI is modeled separately, Sections~4--7 suppress $k$ and write $R_i$. Scores are interpreted within each KPI because reward aggregation, filtering, and scale differ.

\subsection{Focal-Action Conditional Effect Target}
For category $a\in\mathcal A$, let $T_{i,a}$ indicate whether the category is present in the action month and let $R^a(t)$ denote the potential modeling reward under focal exposure $T_a=t$. We define the focal-action conditional average treatment effect as
\begin{equation}
 \tau_a(x)=\mathbb E\left[R^a(1)-R^a(0)\mid X=x\right].
 \label{eq:focal-cate}
\end{equation}
The target intentionally averages over the co-action environment observed with and without the focal category, matching the policy task of ranking one category from bundled historical changes. Under consistency, conditional exchangeability given pre-decision information, and positivity, the corresponding conditional treated-control contrast identifies $\tau_a(x)$ \cite{rosenbaum1983central}. The target motivates selecting the category with the largest estimated focal contrast for the current pre-decision state. A deterministic policy $\pi:\mathcal X\rightarrow\mathcal A$ maps each state to one category, and all candidate policies are compared by the shared test-set score in Section~\ref{sec:score}.

\section{Policies}
\subsection{Contextual Conservative Value Model}
For this baseline, the training data are expanded to one row for each pair $(i,a)$ with $T_{i,a}=1$; $A\in\mathcal A$ denotes the active category attached to an expanded row. Company-months with no active supported category contribute no expanded row. The contextual value model learns $Q_\theta(x,a)$ and selects the largest value over the catalog. Its objective, with expectations over expanded training rows, combines a CQL-style log-sum-exp penalty with regression to the normalized KPI reward:
\begin{equation}
\begin{aligned}
 \min_\theta\quad &\alpha\,\mathbb E\!\left[\log\sum_{a'\in\mathcal A}e^{Q_\theta(X,a')}-Q_\theta(X,A)\right] \\
 &+\frac12\mathbb E\!\left[\bigl(Q_\theta(X,A)-R\bigr)^2\right].
\end{aligned}
\label{eq:cql}
\end{equation}
We use a $[512,512,256]$ MLP, conservative weight $\alpha=0.01$, 30,000 updates, batch size 256, and learning rate $10^{-4}$ with \texttt{d3rlpy} \cite{seno2022d3rlpy}. Each expanded row retains the same state and realized reward, exposing every observed supported category to the conservative objective. Test-set scoring returns to the original, unexpanded rows.

\subsection{Uplift T-Learner}
For each category $a$, the T-Learner fits $\hat\mu^{\mathrm{TL}}_{a,t}(x)\approx\mathbb E[R\mid X=x,T_a=t]$ for $t\in\{0,1\}$ and ranks their conditional prediction difference,
\begin{equation}
 \hat\delta_a(x)=\hat\mu^{\mathrm{TL}}_{a,1}(x)-\hat\mu^{\mathrm{TL}}_{a,0}(x),
 \qquad \pi_{\mathrm{TL}}(x)=\arg\max_{a\in\mathcal A}\hat\delta_a(x).
 \label{eq:tlearner}
\end{equation}
The regressors use only decision-time features and the normalized KPI reward. LightGBM models use 500 boosting rounds, depth 6, shared regularization across actions and KPIs, and validation-set early stopping \cite{ke2017lightgbm}.

\subsection{Covariate-Adjusted Residual Policy Learning}
CAR-PL denotes \emph{Covariate-Adjusted Residual Policy Learning}, an action-wise R-learner with global support shrinkage. Five-fold company-grouped cross-fitting on the training partition estimates a reward nuisance $\hat m(x)=\mathbb E[R\mid X=x]$ and policy-learning propensities $\hat e^{\mathrm{PL}}_a(x)=\Pr(T_a=1\mid X=x)$. We form
\begin{equation}
 \widetilde R_i=R_i-\hat m(X_i),\qquad
 \widetilde T_{i,a}=T_{i,a}-\hat e^{\mathrm{PL}}_a(X_i),
 \label{eq:residuals}
\end{equation}
and fit one effect head per category by
\begin{equation}
 \hat\delta_a(\cdot)=\arg\min_{g\in\mathcal G}\sum_i
 \left(\widetilde R_i-\widetilde T_{i,a}g(X_i)\right)^2.
 \label{eq:rlearner}
\end{equation}
Here $\mathcal G$ is the class of depth-4 gradient-boosted regression heads. We implement Eq.~\eqref{eq:rlearner} as weighted regression: the pseudo-outcome is $\widetilde R_i/\widetilde T_{i,a}$, and the weight is $\widetilde T_{i,a}^2$. Rows with weight below $10^{-4}$ are omitted; each head uses 150 rounds.

Training proceeds in five stages. We first assign firms to five nuisance folds and fit the reward and propensity models on the complementary firms. We then generate out-of-fold residuals for every training row, fit one weighted effect head per category, and evaluate candidate head settings on the company-disjoint validation partition. Finally, the selected support multiplier is applied uniformly across categories before the policy argmax. The action heads share the nuisance residuals but are fit independently and can be trained in parallel, making the method practical for a catalog with dozens of candidate categories.

Maximizing over 34 noisy heads can favor low-frequency categories by chance. Let $n_a=\sum_{i\in\mathcal D_{\mathrm{train}}}T_{i,a}$ be category $a$'s training support. Validation considers $\lambda_{\mathrm{sup}}\in\{0,1000,2000,4000\}$ and yields $\lambda_{\mathrm{sup}}=2000$ for all three KPI policies. CAR-PL then applies the support multiplier
\begin{align}
 \hat\delta_a^{\mathrm{shr}}(x)
 &=\hat\delta_a(x)\frac{n_a}{n_a+\lambda_{\mathrm{sup}}},\quad \lambda_{\mathrm{sup}}=2000,\\
 \pi_{\mathrm{CAR}}(x)
 &=\arg\max_a\hat\delta_a^{\mathrm{shr}}(x).
 \label{eq:shrinkage}
\end{align}
The multiplier favors better-supported categories in the many-category argmax while retaining state-specific variation within each head. CAR-PL operates on the original, unexpanded multi-hot rows; within each action head, rows below the weight threshold are omitted. Because nuisance predictions are out of fold and folds are grouped by company, each effect head is trained on residualized signals from firms not used to fit its nuisance values.

\subsection{Reference Policies}
The zero-shot policy uses \texttt{anthropic.claude-opus-4-6-v1}. It receives the full, untruncated pre-decision narrative (median approximately 23,000 characters), the target KPI, and the 34 catalog names. Its fixed template asks for the single category most likely to improve that KPI and requires the exact category name in JSON. We generate one completion per state with reasoning effort set to \texttt{high}, and every reported output matches an eligible catalog name.

The modal constant policy recommends the most frequent eligible training category for every state. It provides a simple non-personalized comparator. The random reference samples category $a$ once per test row with probability proportional to its training frequency, renormalized over the same catalog. A fixed random seed is used, and the resulting test-row assignments are held constant across all company-clustered bootstrap replicates.

\begin{table*}[t]
\centering
\footnotesize
\setlength{\tabcolsep}{4pt}
\caption{Policy and scoring-model implementation summary. Model and preprocessing choices are selected with development data and frozen before test-set scoring.}
\label{tab:implementation}
\begin{tabular}{p{2.5cm}p{3.1cm}p{3.0cm}p{6.4cm}}
\toprule
Component & Base model & Training target & Main settings \\
\midrule
Contextual value & $[512,512,256]$ Q-MLP & Normalized KPI reward & $\alpha=0.01$, 30k updates, batch 256, learning rate $10^{-4}$ \\
T-Learner & Per-action LightGBM regressors & Normalized KPI reward & 500 rounds, learning rate 0.01, depth 6, early stopping \\
CAR-PL & Weighted GBM R-heads & Residualized normalized KPI reward & Company-grouped 5-fold cross-fitting, 150 rounds, depth 4, $\lambda_{\mathrm{sup}}=2000$ for all KPIs \\
Propensity model & Shared-trunk multi-label MLP & Category indicators & Hidden dimensions 768/256, LayerNorm, ReLU, dropout 0.4, clipping $[0.05,0.95]$ \\
Outcome reference model & Shared 11-head neural model & Monthly post-action KPI values & Pre-decision $W$, multi-hot treatment input, Eq.~\eqref{eq:reward} aggregation, Huber loss, AdamW, early stopping \\
Zero-shot LLM & Claude Opus 4.6 & Direct catalog selection & One completion, reasoning effort \texttt{high}, exact JSON output \\
Modal constant / random & Training frequencies & Non-personalized references & Most frequent category / one fixed seeded frequency-proportional draw \\
\bottomrule
\end{tabular}
\end{table*}

\section{Shared Model-Assisted Policy Comparison}\label{sec:score}
\subsection{Shared Nuisance Models and Reference Contrast}
All policies are compared using the same frozen scorer. Its propensity model uses the policy-state representation $X_i$, while its outcome model uses $W_i$, a structured encoding of KPI histories and accounting covariates from the same pre-decision records. Both representations exclude action-month and post-action information. A shared-trunk MLP estimates scoring propensities $\hat e^{\mathrm{sc}}_a(x)=\Pr(T_a=1\mid X=x)$. The model is fit with company-grouped folds on the training partition; validation data determine early stopping and the frozen configuration. The test-set scoring rule uses $\widetilde e^{\mathrm{sc}}_a(x)=\operatorname{clip}(\hat e^{\mathrm{sc}}_a(x),0.05,0.95)$.

A neural outcome model produces 11 post-action monthly KPI predictions from pre-decision covariates $W$ and multi-hot treatment vector $\mathbf t$. The implemented fixed KPI-specific map aggregates those outputs with the pre-period baseline in $W$ and applies the reward construction in Eq.~\eqref{eq:reward}; $\hat\mu(w,\mathbf t)$ denotes the resulting scalar reward prediction. The network uses Huber loss, AdamW \cite{loshchilov2019decoupled}, and validation-set early stopping. For focal category $a$, we evaluate it at the one-hot vector $\mathbf u_a$ and at the zero vector:
\begin{equation}
 \hat m^{\mathrm{ref}}_{a,1}(w)=\hat\mu(w,\mathbf u_a),\qquad
 \hat m^{\mathrm{ref}}_{a,0}(w)=\hat\mu(w,\mathbf 0).
 \label{eq:anchors}
\end{equation}
These standardized probes place all categories on one model-based reference scale and allow a shared scoring model to compare policies that select different categories.

\subsection{Bounded Augmented Policy Score}
Let $\operatorname{clip}(z,\ell,u)$ truncate $z$ to $[\ell,u]$. For test observation $i$ and focal category $a$, define
\begin{align}
 \hat s_i(a)=&\;\hat m^{\mathrm{ref}}_{a,1}(W_i)-\hat m^{\mathrm{ref}}_{a,0}(W_i) \nonumber\\
 &+\operatorname{clip}\Bigg(
 \frac{T_{i,a}}{\widetilde e^{\mathrm{sc}}_a(X_i)}
 [R_i-\hat m^{\mathrm{ref}}_{a,1}(W_i)] \nonumber\\
 &\qquad-\frac{1-T_{i,a}}{1-\widetilde e^{\mathrm{sc}}_a(X_i)}
 [R_i-\hat m^{\mathrm{ref}}_{a,0}(W_i)],-2,2\Bigg).
 \label{eq:score}
\end{align}
The first term is a shared outcome-model reference contrast, and the bounded residual term incorporates observed focal-exposure evidence while limiting the influence of extreme weights and outcomes. The construction draws on augmented inverse-propensity scoring used in causal and off-policy evaluation \cite{robins1994estimation,dudik2011doubly,thomas2016data}, with the same clipping, reward transformation, and correction bound applied to every policy. Because the outcome-model probes use standardized treatment references rather than each row's factual co-action bundle, we interpret Eq.~\eqref{eq:score} as a common model-assisted score for matched within-KPI policy comparisons, rather than as an estimate of deployed policy value.

The shared scorer is deliberately policy-agnostic. For each test state, it computes action-specific quantities from the same fitted nuisance models and then extracts only the score of the category selected by the candidate policy. Differences between policies therefore arise from their state-to-category mappings rather than from policy-specific scoring models, coverage filters, or outcome transformations.

For the current KPI, let $\mathcal D_{\mathrm{test}}$ be its retained test rows and $N=|\mathcal D_{\mathrm{test}}|$. For any realized one-category mapping $\pi$, the reported metric is the mean selected-category score,
\begin{equation}
 \widehat S_{\mathrm{MA}}(\pi)=\frac{1}{N}\sum_{i\in\mathcal D_{\mathrm{test}}}
 \hat s_i\!\left(\pi(X_i)\right).
 \label{eq:policy-score}
\end{equation}

\subsection{Matched Comparisons and Inference}
Every policy is constrained to $\mathcal A$ before inference. Value-model heads outside the catalog are masked, the T-Learner and CAR-PL take their argmax only over $\mathcal A$, the LLM prompt contains only catalog names, and the random reference renormalizes its probabilities over the same set. Consequently, each policy emits one scorable category for every valid test row; no policy-specific recommendation is dropped after prediction.

All model selection, early stopping, nuisance fitting, action-set construction, reward transformation, and scoring constants use the training and validation partitions. The complete pipeline is frozen before test-set evaluation. Because every policy is scored on the same company-months with the same frozen nuisance models, matched policy differences are the primary comparative quantity. For two policies $\pi_1$ and $\pi_2$, the matched contrast is
\begin{equation}
 \widehat\Delta(\pi_1,\pi_2)=\frac{1}{N}\sum_{i\in\mathcal D_{\mathrm{test}}}
 \left[\hat s_i(\pi_1(X_i))-\hat s_i(\pi_2(X_i))\right].
 \label{eq:paired}
\end{equation}
We form 95\% confidence intervals with 1,000 bootstrap replicates that resample held-out test firms and retain all of each sampled firm's months. Firm-level resampling preserves within-firm dependence, including dependence induced by overlapping monthly outcome windows. For the random reference, the original seeded test-row assignments remain fixed while firms are resampled. The same clustered bootstrap is applied to Eq.~\eqref{eq:policy-score} and Eq.~\eqref{eq:paired}.

\section{Results}
Table~\ref{tab:main-results} reports the shared test-set comparison. CAR-PL and the T-Learner have the two highest point estimates for both growth KPIs, while the contextual value model ranks highest on Quick Ratio. The modal constant has a higher point estimate than the random reference but remains below the highest-scoring learned policy on every KPI.

\begin{table*}[t]
\centering
\small
\setlength{\tabcolsep}{6pt}
\caption{Model-assisted focal-action scores on company-disjoint held-out test firms. Cells show mean [95\% company-clustered bootstrap interval]. Bold marks the highest point estimate. Scores are comparable within, not across, KPIs.}
\label{tab:main-results}
\begin{tabular}{lccc}
\toprule
Policy & Gross Profit & Revenue & Quick Ratio \\
\midrule
CAR-PL & $\mathbf{0.084}\ [0.056,0.110]$ & $0.079\ [0.054,0.102]$ & $0.007\ [-0.032,0.045]$ \\
T-Learner & $0.074\ [0.047,0.100]$ & $\mathbf{0.085}\ [0.061,0.110]$ & $0.018\ [-0.018,0.054]$ \\
Contextual value & $0.048\ [0.020,0.073]$ & $0.059\ [0.034,0.082]$ & $\mathbf{0.062}\ [0.019,0.101]$ \\
Zero-shot LLM & $0.031\ [0.003,0.059]$ & $0.070\ [0.046,0.095]$ & $0.037\ [-0.004,0.078]$ \\
Modal constant & $0.038\ [0.008,0.068]$ & $0.038\ [0.014,0.066]$ & $0.043\ [-0.001,0.087]$ \\
Random reference & $0.022\ [-0.004,0.047]$ & $0.018\ [-0.003,0.037]$ & $0.020\ [-0.016,0.051]$ \\
\bottomrule
\end{tabular}
\end{table*}

\subsection{KPI-Specific Rankings}
On Gross Profit, CAR-PL has the highest point estimate (0.084), followed by the T-Learner (0.074). Their paired difference is 0.010 with a 95\% interval of $[-0.008,0.028]$, so the two policies are not statistically separated. CAR-PL has positive paired margins over the contextual value model (0.037, $[0.016,0.056]$) and random reference (0.063, $[0.041,0.085]$). Revenue yields the same top pair in reverse order: the T-Learner scores 0.085 and CAR-PL 0.079, with a paired difference of $-0.006$ and interval $[-0.025,0.012]$; these policies are again not statistically separated. CAR-PL exceeds the contextual value model and random reference on Revenue. On Quick Ratio, the contextual value model has the highest point estimate (0.062) and positive paired margins over the T-Learner and random reference. Thus, the matched comparisons do not separate CAR-PL and the T-Learner on either growth KPI; on Quick Ratio, the contextual value model has the highest point estimate and positive reported margins over the T-Learner and random reference.

\begin{table}[t]
\centering
\scriptsize
\setlength{\tabcolsep}{3.1pt}
\caption{Selected matched policy-score differences. Intervals use the company-clustered bootstrap. Differences use unrounded row-level scores and may differ slightly from differences between rounded means in Table~\ref{tab:main-results}.}
\label{tab:paired}
\begin{tabular}{llrr}
\toprule
KPI & Contrast & Difference & 95\% CI \\
\midrule
GP & CAR-PL $-$ T-Learner & 0.010 & $[-0.008,0.028]$ \\
GP & CAR-PL $-$ Contextual value & 0.037 & $[0.016,0.056]$ \\
GP & CAR-PL $-$ Random & 0.063 & $[0.041,0.085]$ \\
Revenue & CAR-PL $-$ T-Learner & $-0.006$ & $[-0.025,0.012]$ \\
Revenue & CAR-PL $-$ Contextual value & 0.020 & $[0.002,0.039]$ \\
Revenue & CAR-PL $-$ Random & 0.061 & $[0.044,0.079]$ \\
QR & Contextual value $-$ T-Learner & 0.044 & $[0.015,0.074]$ \\
QR & Contextual value $-$ Random & 0.043 & $[0.010,0.071]$ \\
\bottomrule
\end{tabular}
\end{table}

\subsection{Scoring-Model Stability and Overlap}
Table~\ref{tab:diagnostics} summarizes propensity calibration, overlap, and factual outcome-model diagnostics. Across the 34 categories, macro AUPRC is 0.423/0.428 (training/test) and macro AUROC is 0.819/0.819. Expected calibration error (ECE) is 0.036. The treated-side clip rate is 4.8\%, median inverse-weight effective sample size (ESS) is 58\% of nominal treated support and 93\% of nominal control support, and outcome-model rank correlations are positive on held-out test firms for every KPI.

\begin{table}[t]
\centering
\scriptsize
\setlength{\tabcolsep}{3.8pt}
\caption{Nuisance-model and overlap diagnostics. Propensity AUROC and AUPRC are macro-averaged across the 34 categories. Where three KPI-specific values appear, they are ordered Gross Profit / Revenue / Quick Ratio; other pairs are identified in the row label. Outcome-model diagnostics use held-out test rows.}
\label{tab:diagnostics}
\begin{tabular}{lr}
\toprule
Diagnostic & Value \\
\midrule
Propensity AUROC (train / test) & 0.819 / 0.819 \\
Propensity AUPRC (train / test) & 0.423 / 0.428 \\
Propensity Brier score / ECE & 0.105 / 0.036 \\
Treated propensity clip rate & 4.8\% \\
Median treated/control ESS & 58\% / 93\% \\
Outcome-model MAE & 0.402 / 0.331 / 0.652 \\
Outcome-model Spearman $\rho$ & 0.213 / 0.238 / 0.383 \\
Outcome residual IQR & 0.439 / 0.382 / 0.779 \\
\bottomrule
\end{tabular}
\end{table}

Table~\ref{tab:saturation} gives the fraction of test rows whose residual correction reaches $\pm2$. Saturation is 2.6--5.5\% on the growth KPIs and 6.6--12.9\% on Quick Ratio, so the correction bound is inactive on most rows.

\begin{table}[t]
\centering
\scriptsize
\setlength{\tabcolsep}{4.4pt}
\caption{Residual-correction saturation rate at the headline score settings.}
\label{tab:saturation}
\begin{tabular}{lrrr}
\toprule
Policy & Gross Profit & Revenue & Quick Ratio \\
\midrule
CAR-PL & 3.5\% & 2.6\% & 10.8\% \\
T-Learner & 5.0\% & 3.8\% & 12.9\% \\
Contextual value & 5.4\% & 3.7\% & 6.6\% \\
Zero-shot LLM & 5.5\% & 4.0\% & 12.3\% \\
\bottomrule
\end{tabular}
\end{table}

Table~\ref{tab:sensitivity} reports the Gross Profit CAR-PL grid. Propensity bounds are 0.01, 0.05, and 0.10; correction bounds are 1, 2, 5, and unbounded. Every score remains positive, ranging from 0.083 to 0.118. On Revenue, CAR-PL ranges from 0.077 to 0.095, and the contextual value model's Quick Ratio score remains between 0.06 and 0.10. We additionally report an outcome-model-only score that omits the residual correction in Eq.~\eqref{eq:score}. Separately, we replace the all-zero treatment reference in Eq.~\eqref{eq:anchors} with the modal training co-action vector, defined as the most common treatment configuration in the training data. Table~\ref{tab:alt-score} shows that the outcome-model-only analysis retains the same KPI-level point-estimate leader or top pair. Category-contrast rankings under the two treatment references are strongly correlated ($\rho=0.884$--$0.971$), and the top-ranked category is unchanged for every KPI.

\begin{table}[t]
\centering
\scriptsize
\setlength{\tabcolsep}{4.1pt}
\caption{Gross Profit CAR-PL score under propensity and residual-correction bounds. The headline setting is $(0.05,2)$.}
\label{tab:sensitivity}
\begin{tabular}{lrrrr}
\toprule
Propensity bound & Corr. 1 & Corr. 2 & Corr. 5 & Unbounded \\
\midrule
0.01 & 0.103 & 0.083 & 0.084 & 0.118 \\
0.05 & 0.105 & \textbf{0.084} & 0.087 & 0.107 \\
0.10 & 0.110 & 0.087 & 0.093 & 0.103 \\
\bottomrule
\end{tabular}
\end{table}

\begin{table}[t]
\centering
\scriptsize
\setlength{\tabcolsep}{2.6pt}
\caption{Robustness to score component and treatment reference. Outcome-model-only leader(s) use only the reference contrast in Eq.~\eqref{eq:score}. Zero/modal $\rho$ is the Spearman correlation across 34 category contrasts under the all-zero and modal co-action treatment references.}
\label{tab:alt-score}
\begin{tabular}{lp{2.5cm}p{1.35cm}p{1.25cm}}
\toprule
KPI & Outcome-model-only leader(s) & \shortstack{Zero/\\modal $\rho$} & Top unchanged \\
\midrule
Gross Profit & T-Learner 0.039; CAR-PL 0.039 & 0.971 & Yes \\
Revenue & T-Learner 0.085; CAR-PL 0.069 & 0.884 & Yes \\
Quick Ratio & Contextual value 0.027 & 0.971 & Yes \\
\bottomrule
\end{tabular}
\end{table}

\subsection{Policy Behavior and Constant References}
Table~\ref{tab:behavior} describes test-set selection concentration. CAR-PL uses 33--34 categories with top-category shares below 19\% on all three KPIs. The T-Learner is more concentrated on the two growth KPIs, while the zero-shot LLM assigns 84.2\% of its Revenue recommendations to one broad customer-growth category. CAR-PL and the T-Learner choose the same category on only 12.8\% of Gross Profit states, 14.2\% of Revenue states, and 10.7\% of Quick Ratio states. This broader catalog use indicates that CAR-PL does not obtain its score by collapsing onto a small set of generic categories and produces broader, less concentrated selections across the action catalog.

\begin{table}[t]
\centering
\scriptsize
\setlength{\tabcolsep}{3.3pt}
\caption{Test-set action-selection concentration. Entropy is $-\sum_a p_a\log_2 p_a$ bits for empirical selection shares $p_a$.}
\label{tab:behavior}
\begin{tabular}{llrrrr}
\toprule
KPI & Policy & Distinct & Top-1 & Top-5 & Entropy \\
\midrule
GP & CAR-PL & 34 & 13.8\% & 46.7\% & 4.39 \\
GP & T-Learner & 22 & 25.2\% & 61.8\% & 3.68 \\
GP & LLM & 18 & 39.5\% & 93.6\% & 2.48 \\
Revenue & CAR-PL & 33 & 18.7\% & 51.0\% & 4.19 \\
Revenue & T-Learner & 22 & 30.1\% & 66.1\% & 3.50 \\
Revenue & LLM & 20 & 84.2\% & 94.7\% & 1.13 \\
QR & CAR-PL & 34 & 8.0\% & 29.9\% & 4.87 \\
QR & Contextual value & 26 & 37.3\% & 91.9\% & 2.39 \\
\bottomrule
\end{tabular}
\end{table}

The modal constant provides a simple non-personalized comparator. Its scores are 0.038 on Gross Profit, 0.038 on Revenue, and 0.043 on Quick Ratio, below the highest-scoring learned policy on every KPI. Together with the concentration results, this shows that the highest-scoring learned mappings use a substantially more diverse set of categories than the one-category modal rule.

\section{Discussion}
\subsection{KPI-Dependent Results}
Quick Ratio presents a noisier residualized learning setting than the growth KPIs: its outcome-residual IQR is 0.779 versus 0.382 for Revenue, and its CAR-PL correction saturation is 10.8\% versus 2.6\%. This pattern may partly explain why the contextual value model has the highest Quick Ratio point estimate.

On Gross Profit and Revenue, CAR-PL and the T-Learner are not statistically separated in matched comparisons while agreeing on only 12.8\% and 14.2\% of states, respectively. Comparable aggregate scores can therefore arise from substantially different policy mappings, making recommendation coverage and concentration important complements to the mean policy score. Matched comparisons further show that CAR-PL exceeds the contextual value model on the growth KPIs, whereas the contextual value model exceeds the T-Learner and random reference on Quick Ratio.

\subsection{CAR-PL in the Comparison}
CAR-PL's support-shrunk action-wise residual estimates directly address instability from selecting among 34 unevenly supported categories. On the growth KPIs, CAR-PL is not statistically separated from the T-Learner while selecting 33--34 categories and maintaining top-category shares below 19\%. It therefore combines competitive growth-KPI performance with broader recommendation coverage across heterogeneous firm states and business domains.

\subsection{Learning from Multi-Action Accounting Logs}
The multi-hot structure is central to the study. A company may change spending, financing, vendors, and operating tools in the same month, so the learning methods must extract a useful focal signal from repeated bundled observations. The contextual value model converts each active category into a discrete training row. The T-Learner and CAR-PL retain the original rows and fit one focal category at a time, with CAR-PL additionally residualizing both outcome and category occurrence against the decision state. These constructions provide complementary ways to exploit the same logs.

The treatment-reference analysis provides a complementary check on learning from bundled logs. Per-category outcome-model contrasts remain strongly rank-aligned when the all-zero treatment reference is replaced by the modal training co-action vector, and the highest-ranked category is unchanged for every reported KPI. Together with the outcome-model-only comparison, this indicates that the headline KPI-level point-estimate leader or top pair is not specific to the residual correction or to one treatment reference.

\subsection{Reference Policies and Practical Interpretation}
The zero-shot LLM is competitive on Revenue, but its 84.2\% concentration on one category indicates that generic business priors produce a much narrower policy than outcome-fitted models. The comparison distinguishes broad pretrained knowledge from state-specific policy learning. The modal constant and frequency-weighted random references are both below the highest-scoring learned policy on every KPI, a pattern consistent with outcome fitting adding useful information beyond the historical marginal action distribution. The single-category output also fits a human-review workflow: a policy surfaces one prioritized business-change category, while an advisor can translate it into a firm-specific implementation plan.

\subsection{Evaluation Design}
Four design choices support the comparison. Every method shares the 34-category action space and frozen scorer. Development and testing are company-disjoint, pairwise claims use matched differences on the same company-months, and separate policies are learned for each financial objective. This isolates state-to-category mappings from coverage and scorer changes while respecting the different scales and business meanings of Gross Profit, Revenue, and Quick Ratio. The accounting records and exact detector thresholds are proprietary; the paper therefore reports the partitioning design, prompt configuration, model architectures, hyperparameters, score constants, and aggregate diagnostics to support reimplementation on an equivalent corpus.

\subsection{Limitations}
The study uses observational accounting data and therefore relies on measured pre-decision covariates, overlap, and the quality of ledger-derived category labels. Co-occurring business changes are represented through focal categories rather than explicit bundle policies, and the model-assisted score depends on fitted nuisance models and standardized treatment references. The company-disjoint test assesses generalization within the observed environment rather than the effect of deployed recommendations. Prospective validation is the next step toward operational deployment.

\section{Conclusion}
We presented an observational policy-ranking framework for SMB financial guidance from multi-action accounting logs. Across 85,078 company-month observations, CAR-PL and the T-Learner have the two highest point estimates and are not statistically separated on Gross Profit or Revenue, while the contextual value model has the highest Quick Ratio point estimate. CAR-PL combines competitive growth-KPI performance with broader recommendation coverage, selecting 33--34 categories across held-out firms. Outcome-model-only scoring retains the same KPI-level point-estimate leader or top pair, and category rankings remain similar when the all-zero treatment reference is replaced by the modal training co-action vector. Overall, the findings support objective-specific comparison of learning strategies and position CAR-PL as a practical support-aware method for ranking guidance over large observational action catalogs.

\balance
\bibliographystyle{unsrtnat}
\bibliography{references}

\begin{thebibliography}{22}
\providecommand{\natexlab}[1]{#1}
\providecommand{\url}[1]{\texttt{#1}}
\expandafter\ifx\csname urlstyle\endcsname\relax
  \providecommand{\doi}[1]{doi: #1}\else
  \providecommand{\doi}{doi: \begingroup \urlstyle{rm}\Url}\fi

\bibitem[Bruhn et~al.(2018)Bruhn, Karlan, and Schoar]{bruhn2018}
Miriam Bruhn, Dean Karlan, and Antoinette Schoar.
\newblock The impact of consulting services on small and medium enterprises:
  Evidence from a randomized trial in mexico.
\newblock \emph{Journal of Political Economy}, 126\penalty0 (2):\penalty0
  635--687, 2018.
\newblock \doi{10.1086/696154}.

\bibitem[Athey and Wager(2021)]{athey2021policy}
Susan Athey and Stefan Wager.
\newblock Policy learning with observational data.
\newblock \emph{Econometrica}, 89\penalty0 (1):\penalty0 133--161, 2021.
\newblock \doi{10.3982/ECTA15732}.

\bibitem[Zhou et~al.(2023)Zhou, Athey, and Wager]{zhou2018offline}
Zhengyuan Zhou, Susan Athey, and Stefan Wager.
\newblock Offline multi-action policy learning: Generalization and
  optimization.
\newblock \emph{Operations Research}, 71\penalty0 (1):\penalty0 148--183, 2023.
\newblock \doi{10.1287/opre.2022.2271}.

\bibitem[VanderWeele and Hern{\'a}n(2013)]{vanderweele2013versions}
Tyler~J. VanderWeele and Miguel~A. Hern{\'a}n.
\newblock Causal inference under multiple versions of treatment.
\newblock \emph{Journal of Causal Inference}, 1\penalty0 (1):\penalty0 1--20,
  2013.
\newblock \doi{10.1515/jci-2012-0002}.

\bibitem[K{\"u}nzel et~al.(2019)K{\"u}nzel, Sekhon, Bickel, and
  Yu]{kunzel2019metalearners}
S{\"o}ren~R. K{\"u}nzel, Jasjeet~S. Sekhon, Peter~J. Bickel, and Bin Yu.
\newblock Metalearners for estimating heterogeneous treatment effects using
  machine learning.
\newblock \emph{Proceedings of the National Academy of Sciences}, 116\penalty0
  (10):\penalty0 4156--4165, 2019.
\newblock \doi{10.1073/pnas.1804597116}.

\bibitem[Nie and Wager(2021)]{nie2021quasi}
Xinkun Nie and Stefan Wager.
\newblock Quasi-oracle estimation of heterogeneous treatment effects.
\newblock \emph{Biometrika}, 108\penalty0 (2):\penalty0 299--319, 2021.
\newblock \doi{10.1093/biomet/asaa076}.

\bibitem[Jin et~al.(2025)Jin, Ren, Yang, and Wang]{jin2025pessimism}
Ying Jin, Zhimei Ren, Zhuoran Yang, and Zhaoran Wang.
\newblock Policy learning ``without'' overlap: Pessimism and generalized
  empirical bernstein's inequality.
\newblock \emph{The Annals of Statistics}, 53\penalty0 (4):\penalty0
  1483--1512, 2025.
\newblock \doi{10.1214/25-AOS2511}.

\bibitem[Fujimoto et~al.(2019)Fujimoto, Meger, and Precup]{fujimoto2019bcq}
Scott Fujimoto, David Meger, and Doina Precup.
\newblock Off-policy deep reinforcement learning without exploration.
\newblock In \emph{Proceedings of the 36th International Conference on Machine
  Learning}, volume~97 of \emph{Proceedings of Machine Learning Research},
  pages 2052--2062, 2019.

\bibitem[Levine et~al.(2020)Levine, Kumar, Tucker, and Fu]{levine2020offline}
Sergey Levine, Aviral Kumar, George Tucker, and Justin Fu.
\newblock Offline reinforcement learning: Tutorial, review, and perspectives on
  open problems.
\newblock \emph{arXiv preprint arXiv:2005.01643}, 2020.

\bibitem[Kumar et~al.(2020)Kumar, Zhou, Tucker, and Levine]{kumar2020cql}
Aviral Kumar, Aurick Zhou, George Tucker, and Sergey Levine.
\newblock Conservative {Q}-learning for offline reinforcement learning.
\newblock In \emph{Advances in Neural Information Processing Systems},
  volume~33, pages 1179--1191, 2020.

\bibitem[Khandani et~al.(2010)Khandani, Kim, and Lo]{khandani2010consumer}
Amir~E. Khandani, Adlar~J. Kim, and Andrew~W. Lo.
\newblock Consumer credit-risk models via machine-learning algorithms.
\newblock \emph{Journal of Banking \& Finance}, 34\penalty0 (11):\penalty0
  2767--2787, 2010.
\newblock \doi{10.1016/j.jbankfin.2010.06.001}.

\bibitem[Fuster et~al.(2022)Fuster, Goldsmith-Pinkham, Ramadorai, and
  Walther]{fuster2022predictably}
Andreas Fuster, Paul Goldsmith-Pinkham, Tarun Ramadorai, and Ansgar Walther.
\newblock Predictably unequal? the effects of machine learning on credit
  markets.
\newblock \emph{The Journal of Finance}, 77\penalty0 (1):\penalty0 5--47, 2022.
\newblock \doi{10.1111/jofi.13090}.

\bibitem[Dai et~al.(2023)Dai, Shao, Zhao, Yu, Si, Xu, Sun, Zhang, and
  Xu]{dai2023llm_rec}
Sunhao Dai, Ninglu Shao, Haiyuan Zhao, Weijie Yu, Zihua Si, Chen Xu, Zhongxiang
  Sun, Xiao Zhang, and Jun Xu.
\newblock Uncovering {ChatGPT}'s capabilities in recommender systems.
\newblock \emph{arXiv preprint arXiv:2305.02182}, 2023.

\bibitem[Fieberg et~al.(2025)Fieberg, Hornuf, Meiler, and
  Streich]{fieberg2025llm_finance}
Christian Fieberg, Lars Hornuf, Maximilian Meiler, and David~J. Streich.
\newblock Using large language models for financial advice.
\newblock CESifo Working Paper 11666, CESifo, 2025.

\bibitem[Zhang et~al.(2025)Zhang, Li, Long, Zhang, Lin, Yang, Xie, Yang, Liu,
  Lin, Huang, and Zhou]{qwen3embedding2025}
Yanzhao Zhang, Mingxin Li, Dingkun Long, Xin Zhang, Huan Lin, Baosong Yang,
  Pengjun Xie, An~Yang, Dayiheng Liu, Junyang Lin, Fei Huang, and Jingren Zhou.
\newblock {Qwen3 Embedding}: Advancing text embedding and reranking through
  foundation models.
\newblock \emph{arXiv preprint arXiv:2506.05176}, 2025.

\bibitem[Rosenbaum and Rubin(1983)]{rosenbaum1983central}
Paul~R. Rosenbaum and Donald~B. Rubin.
\newblock The central role of the propensity score in observational studies for
  causal effects.
\newblock \emph{Biometrika}, 70\penalty0 (1):\penalty0 41--55, 1983.
\newblock \doi{10.1093/biomet/70.1.41}.

\bibitem[Seno and Imai(2022)]{seno2022d3rlpy}
Takuma Seno and Michita Imai.
\newblock d3rlpy: An offline deep reinforcement learning library.
\newblock \emph{Journal of Machine Learning Research}, 23\penalty0
  (315):\penalty0 1--20, 2022.

\bibitem[Ke et~al.(2017)Ke, Meng, Finley, Wang, Chen, Ma, Ye, and
  Liu]{ke2017lightgbm}
Guolin Ke, Qi~Meng, Thomas Finley, Taifeng Wang, Wei Chen, Weidong Ma, Qiwei
  Ye, and Tie-Yan Liu.
\newblock {LightGBM}: A highly efficient gradient boosting decision tree.
\newblock In \emph{Advances in Neural Information Processing Systems},
  volume~30, pages 3146--3154, 2017.

\bibitem[Loshchilov and Hutter(2019)]{loshchilov2019decoupled}
Ilya Loshchilov and Frank Hutter.
\newblock Decoupled weight decay regularization.
\newblock In \emph{International Conference on Learning Representations}, 2019.

\bibitem[Robins et~al.(1994)Robins, Rotnitzky, and Zhao]{robins1994estimation}
James~M. Robins, Andrea Rotnitzky, and Lue~Ping Zhao.
\newblock Estimation of regression coefficients when some regressors are not
  always observed.
\newblock \emph{Journal of the American Statistical Association}, 89\penalty0
  (427):\penalty0 846--866, 1994.
\newblock \doi{10.1080/01621459.1994.10476818}.

\bibitem[Dud{\'i}k et~al.(2011)Dud{\'i}k, Langford, and Li]{dudik2011doubly}
Miroslav Dud{\'i}k, John Langford, and Lihong Li.
\newblock Doubly robust policy evaluation and learning.
\newblock In \emph{Proceedings of the 28th International Conference on Machine
  Learning}, pages 1097--1104, 2011.

\bibitem[Thomas and Brunskill(2016)]{thomas2016data}
Philip~S. Thomas and Emma Brunskill.
\newblock Data-efficient off-policy policy evaluation for reinforcement
  learning.
\newblock In \emph{Proceedings of the 33rd International Conference on Machine
  Learning}, volume~48 of \emph{Proceedings of Machine Learning Research},
  pages 2139--2148, 2016.

\end{thebibliography}
\end{document}